\documentclass[runningheads]{llncs}
\usepackage{graphicx}
\usepackage{svg}
\usepackage{url}
\usepackage{comment}
\usepackage{booktabs} 
\usepackage{diagbox}

\usepackage{multirow}
\begin{document}
\title{Applying machine learning to predict behavior of bus transport in Warsaw, Poland}


\author{Łukasz Pałys\inst{1} \and
Maria Ganzha\inst{1}\orcidID{0000-0001-7714-4844} \and
Marcin Paprzycki\inst{2}\orcidID{0000-0002-8069-2152}}
\authorrunning{Ł. Pałys et al.}
%
\institute{Warsaw University of Technology, Warsaw, Poland \\
\email{maria.ganzha@pw.edu.pl}\\
\and
Systems Research Institute Polish Academy of Sciences\\
\email{marcin.paprzycki@ibspan.waw.pl}}
\maketitle   
\begin{abstract}
Nowadays, it is possible to collect precise data describing movements of public transport. Specifically, for each bus (or tram) geoposition data can be regularly collected. This includes data for all buses in Warsaw, Poland. Moreover, this data can be downloaded and analysed. In this context, one of the simplest questions is: can a model be build to represent behavior of busses, and predict their delays. This work provides initial results of our attempt to answer this question.
\end{abstract}
	
\section{Introduction}\label{sec:intro}
One of key aspects of development of Smart Cities is the growing number of installed \textit{sensors}. To make city really smart, streams of data generated by city-sensors are made open to analysis. Moreover, sometimes, privately owned sensors open data to contribute to the well-being of city inhabitants.

Following this trend, Warsaw, Poland, opened its data resources. Here, available data includes, among others, culture, education and urban transport. For instance, data about location of each city bus is regularly collected. The question that arises is: how this data can be used to deliver service to people using public transport? To answer this question we have applied machine learning to predict delays in public transport. Note that such solution (1) could improve effectiveness of movement by means of public transport. Currently, it often happens that public transport vehicles (in particular, buses) arrive inconsistently with their timetable. A delay prediction system (2) would save time spent waiting for a late bus(es). It would also allow for a (3) better planning of journeys requiring a change, taking into account the risk of a late arrival at the bus change location. Another potential advantage of the proposed solution is (4) possibility of predicting ``typical delays'' and, on their basis, modify the timetables. Another way of using the results could be to (5) modify traffic organization, e.g. by introducing bus-passes, thus helping busses to smoothly pass through problematic locations. Moreover, (6) from the results it may be possible to obtain information, which factors influence the delays to the greatest extent. It may be also possible (7) to establish which factors can be successfully captured in machine learning models, and which factors would require special treatment and may be difficult to incorporate. As a result, in the future, (8) it may be possible to undertake actions aiming at removing, or reducing, some causes of delays.

Reported work is based on actual data provided by the city of Warsaw. This data contains bus schedules and their geolocation. Note that tram information is also available. However, we have decided to start our explorations from bus lines. Since our results are promising, we plan to extend our investigations further.

\section{Related work} \label{SotA}
Predicting urban bus travel time has been discussed in the literature. Applied methods can be divided into several groups. First, the statistical methods, such as kNN~\cite{knn2012,ErlendDahl2014}, regression~\cite{ErlendDahl2014}, prediction based on sequence patterns~\cite{PSbasedF2020} or time series~\cite{STL2020}, and application of Kalman filter~\cite{Yang2005}. Second, methods based on observations, which include the historical means approach~\cite{WeiFan,Zychowski2017}. Next, methods based on machine learning, 
which include: back propagation neural network~\cite{BP2019}, radial basis function network~\cite{RBF2014}, multilayer perceptron~\cite{Zychowski2017,WeiFan}. The last group consists of hybrid methods that combine multiple algorithms into a single model~\cite{YongjieLin2013,Hybrid2020}. Let us now discuss the selected methods in more detail.
	
The Pattern Sequence-based Forecasting technique was used in~\cite{PSbasedF2020}, to predict bus travel times in Chennai, India. Discrete Wavelet Transform, and two clustering algorithms were used to extract travel time trends. Results showed that predictions based on the k-means algorithm were more accurate. Predictions have been tested on sections with low, medium and high travel time variability. For sections with low and medium travel time variance, the results were satisfactory. However, for sections with large time differences, they were not. 

The time series approach was used in~\cite{STL2020} to forecast bus travel time in Lviv, Ukraine. The authors used STL (Seasonal and Trend decomposition using Loess) time series decomposition to identify patterns. Travel time data, from~(5) working days, from September-October 2018 was collected for a single bus line, in both directions. The training set consisted of~9 weeks, while test data was from week~10. Systematic variability of traffic intensity was noticed. For travel towards the city center, the highest congestion occurred in the morning, due to commuting to work. For travel from the city, the greatest crowd was observed in the evening. The predictive model obtained a mean absolute percentage error (MAPE) of less than 8\%. The average error, of predicting travel time towards city center was approximately 3 minutes, and 2 minutes in the opposite direction.

In~\cite{RBF2014}, authors used Radial Basis Function Neural Network (RBFNN) to predict bus travel times in Dalian, China. The model was trained on historical data, with additional correction of results. The input data of the model was: time of arrival at stop ``i'', distance between stops ``i'' and ``i + 1'', number of passengers leaving/entering at stop ``i'', delay, average speed between stops ``i'' and ``i + 1''. Correction was based on estimating travel time between consecutive stops, using Kalman Filter. Obtained results were compared with use of BPNN and RBFN networks, without data correction. Overall, the RBFNN model, with online tuning, was the best, with the MAPE at 7.59\%. In comparison, the MAPE values of the non-tuned BPNN and RBFNN models were 17.41\% and 15.98\%.

Bus travel time prediction, described in~\cite{Hybrid2020}, is based on deep learning and data fusion. Here, travel time was divided into driving time and stopping time. Bus GPS data was applied. The models used fourteen features, including: stop ID, day of the week, time, current speed (based on GPS), stopping time at a stop, travel time between stops. Data used for testing came from Guangzhou and Shenzhen (China) and represented a single line in each city. 
To avoid one-hot encoding\footnote{\url{https://en.wikipedia.org/wiki/One-hot}}, they used 
embedding was used for for: time of day, day of the week, stop ID. Moreover, since the size of the ``discrete features'' was much smaller than that of the continuous ones, data was unbalanced. Hence, the Wide and Deep model was used. The Wide component was responsible for the embedded discrete data, while the Deep component analyzed the continuous data using the Deep Neural Network (DNN), and results of both components were concatenated into a feature list. Next, using the Recurrent Neural Network (RNN), the feature weights were determined. Travel time was predicted using DNN. Prediction results were compared with historical averages. Proposed model outperformed other approaches with MAPE at 8.43\% (Gangzhou, line 261). Moreover, MAPE of predictions for the peak hours was 4-7\% lower than they were 9.51\% (for morning hours) and 6.43\% (for evening hours).
In~\cite{WeiFan}, dynamic travel time prediction model was developed, to predict approximate arrival times at subsequent stops. GPS data, from November, 2008 to May, 2009 from Macea (Brazil) for single bus line with 35 stops was used. Here, \textit{Historical Average}, \textit{Kalman Filtering} and \textit{Artificial Neural Network} were tried. Authors discovered considerable variation in average travel time at different times of the day. For the northern direction, in rush hours (17:00-18:00), travel time was 30\% longer than the average time. For the south direction, the mean time during peak hours (7:00-7:30) was about 20\% longer. This fact was taken into account when developing the model. However, Ducan's multiple range test showed that there were no significant differences in travel time distributions in both directions.
The 3-layer perceptron achieved the best result, with overall MAPE at 18.3\%. Importantly, the article shows that bus travel times can be reasonably predicted using only arrival and departure times at stops.

In~\cite{ErlendDahl2014}, the review of methods of predicting bus arrival times, found in~\cite{ErlendDahl2013}, was extended. Authors used: k-Nearest Neighbor, Artificial Neural Network, and Super Vector Regression to predict bus travel times in Trondheim (Norway). Used data included: the route of the vehicle, planned and actual time of arrival, and stopping time. To check the influence of individual time intervals, 423 different data sets were created. To evaluate model performance, mean absolute error (MAE) was used. Depending on data set, (best) MAE varied between 61 and 86 seconds. Moreover, each method delivered a ``winner'' (best result for a data set).
%
%
Separately, additional attributes, e.g.: weather, football matches, tickets, were tried. However, this information did not bring significant improvements. The ticket data improved the results in some training periods of the ANN and KNN models.The weather impact on foreseen methods also have been researched and authors discovered that using travel time instead of delay time will provide more accurate and stable results. The result of the study was the confirmation of this thesis.

Comparing the described classifiers, the authors found that the best model for real-time prediction is kNN. Together with SVR, it achieved the best results, it's training time was by far the shortest and forecasts were the most stable. The analysis of the data sets allowed the authors to conclude that the best way to predict travel time on a given day is to train on the previous three days of the same type. This means that if the goal is to predict the coming Friday, the model should be trained on the last three fives.

In~\cite{YongjieLin2013}, an ANN was used with historical GPS data and an automatic toll collection system data. Moreover, impact of intersections with traffic lights was taken into account. Data came from Jinan (China), bus line 63, and was collected for 30 days from November 1, 2010. In the model, travel time and time distances between successive buses were used. The distribution of time intervals between buses on weekdays and at weekends was analyzed. It was found that these values fluctuate significantly during weekdays, while during weekends the distribution was more stable. Moreover, the average travel time and travel time variances for the following days of the week were determined. Initially an ANN with single hidden layer was used. 
However, due to travel time fluctuations, finally, a hybrid ANN (HANN) model was used. It consists of separate subnets, trained in real-time on different subsets of data, to predict of travel time within specific period, depending on the days (working days, weekends) and peak hours. 
Comparison of prediction results using the relative prediction error (RPE) index demonstrated that the ANN and HANN models are more reliable than the Kalman Filter. Moreover, the HANN model is better suited for short-distance prediction. 

A comparison of the methods of predicting trams travel time in Warsaw, 
is presented in~\cite{Zychowski2017}. Two data sources were used: (1)~tram timetables, including locations of stops and routes, and planned times of arrivals; and (2)~the GPS-based positions of trams. 
Authors tested: delay propagation, historical average (HA), and an ANN model. 
It was found that, overall, the HA was the best method. However, ANN was best in predicting delays in rush hours.


Overall, the main findings from the analysed literature can be summarized as follows. (a)~\textbf{Reported results} have been focused mostly on a single city. Only in one case two different cities have been approached. In all cases a single bus line was studied. Hence, there is no ``benchmark'' data that performance of proposed approaches can be assessed against. (b)~The \textbf{main methods} used to predict bus travel time (delays) were: (i)~statistical methods -- k-nearest neighbours (kNN), regression model, Kalman filter, (ii)~historical observation methods (HA), (iii)~machine learning methods -- back propagation neural networks, radial basis function networks, multilayer perceptrons, and (iv) hybrid methods combining the above algorithms into one model (HANN). Out of these methods, the best results have been reported for HANN, HA, RBFN and multilayer perceptron. However, it was also found that use of HANN requires substantially larger set of training data. (c) \textbf{Quality of prediction}s has been measured using: (i)~mean absolute error (MAE; in seconds); (ii)~mean percentage absolute error (MAPE); (iii)~standard deviation (STD; in seconds). (d) The  \textbf{simplest approaches} used GPS data alone. Other popular data elements were: information about sold tickets and about bus speed. Additionally, effects of non-travel events (e.g. games or weather) have been (unsuccessfully) tried. (e)\textbf{ Best accuracy} was reported in~\cite{ErlendDahl2014}, where MAE was 40s. However, predictions were for 1 to 16 stops only. A good accuracy for a longer bus line was with MAE of the order of 70. However, for the same line, the kNN model performed better, with MAE at 61s. On the other hand, for long Warsaw tram lines (more than 40 stops), the MAE was at 148s. However, in this case the HA reduced MAE to 123s.

Based on these findings, the approach and the main contributions of this work can be summarized as follows. (A)~The MLP, RBFN and HA models were selected. (B)~While still using data from a single city, bus lines, with different characteristics (e.g. normal, express, cross-town, long rout, short rout, etc.) have been studied. (C)~Performance of different approaches to predicting travel time at subsequent stops has been evaluated. (D)~Influence of additional features (data from timetables, number of intersections with traffic lights, information whether the distance between stops includes a bus lane, weather) on the prediction accuracy has been evaluated. (E) Hybrid model, composed of models predicting travel time for various distances, has been proposed.

\section{Available data, its preprocessing, and experimental setup}\label{data}
Let us now describe the data set, and applied preprocesseing. As a part of the project \textit{Open data in Warsaw}\footnote{\url{https://api.um.warszawa.pl/}}, data that includes the exact location of public transport vehicles, reported in real time, is available. From there, 30 days worth of data about busses was harvested. On weekdays, approximately 210MB of data were downloaded. On Saturdays and Sundays, the data was approximately 160 MB. Harvesting lasted from March 8, 2021 till May 10, collecting about 10 GB of data.
Specifically, in order to download the data, a script was written, which queried the API every 30 seconds, to retrieve locations of all buses in use. In response, JSON with a list of busses and their exact locations was received.
A single CSV record contained: (1)~line number, (2)~vehicle ID, (3)~brigade number, (4)~time, (5)~geographic coordinates of the location:
%
%
Based on analysis, presented in Section~\ref{SotA}, it has been decided to use inputs similar to these reported in the literature. In this way, while direct comparison of results is impossible, an indirect comparison will be attempted. Hence, preprocessing was applied, resulting in records containing: (1)~line number, (2)~departure time from the last stop, (3)~current percentage of distance traveled between adjacent stops, (4)~time of the last GPS signal, (5)~current time, (6)~driving direction. Here, use of time of arrival at the stop was based on~\cite{Zychowski2017}. 

\begin{figure}[htb]
    \centering
    \includegraphics[scale=0.3]{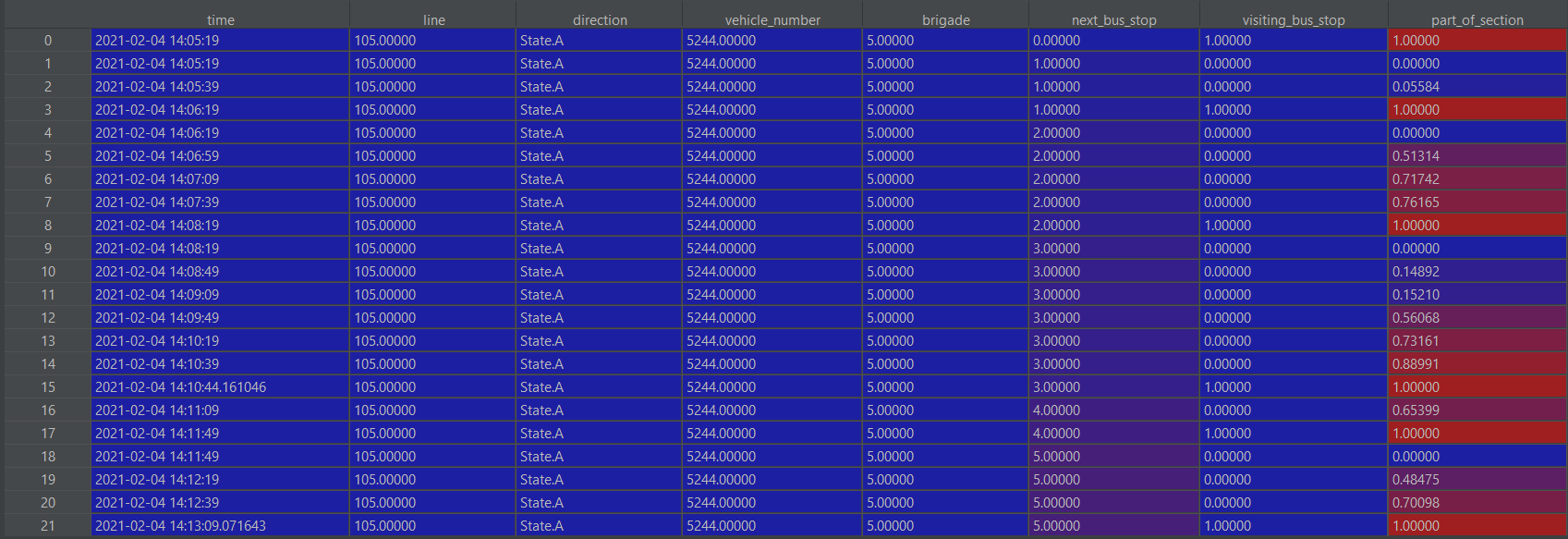}
    \caption{Preprocessed data sample (A/B indicate different directions of the same route}
    \label{fig:my_label1}
\end{figure}

Additionally, file containing timetable of buses, their routes, including list and GPS coordinates of stops, and departure times from individual stops, was downloaded from the ZTM website\footnote{\url{https://www.wtp.waw.pl}}. A special script has been prepared to extract all necessary data from this file. 
Pre-processing of this file created two tables, containing: (1)~geographical coordinates of all stops, and (2)~lists of all stops for a given line and direction. 
After preprocessing, file with: line number, time, vehicle and brigade number,driving direction, number of the next stop to visit on the route, information whether the vehicle is at the stop, the percentage of distance traveled between consecutive stops, was created (see, Figure~\ref{fig:my_label1}).

Separately, file with additional features has been prepared, containing: section congestion, timetable, and weather information. The congestion captures number of busses traveling at a given moment the route between neighboring stops. While in~\cite{ErlendDahl2014} it was claimed that weather information did not significantly improve the prediction quality, it was decided to test this for Warsaw busses. 

From available data, 29 bus lines, that sent an average of more than 22,000 GPS signals a day, were selected. Based on careful analysis of characteristics of their routes, within the context of the city itself, selected bus lines have been manually grouped into eight groups with similar characteristics: 
\begin{enumerate}
    \item Long routes within periphery North-South: bypassing the City Center, running on the western side of the Vistula River; lines: 136, 154, 167, 187, 189.
    \item Long routes within periphery West-East: bypassing the City Center, crossing the Vistula River; lines 112, 186, 523.
    \item Centre-periphery: routes with one end in the City Center, running to the peripheries; not crossing the river; lines 131, 503, 504, 517, 518.
    \item Long routes through Center (with ends on peripheries); lines 116, 180, 190.
    \item Express: a fairly straight long lines with small number of stops (typically around 1/3 stops of ``normal'' lines); lines 158, 521, 182, 509.
    \item Centre-Praga: short routes starting in the City Center; crossing the Vistula River; lines 111, 117, 102.
    \item Short lines within peripheries: located in Western part of Warsaw; lines 172, 191, 105.
    \item Short lines within the Center: not crossing the river; lines 128, 107, 106.
\end{enumerate}

The base \textit{MLP} and \textit{RBFN} models described in~\cite{Zychowski2017} and~\cite{RBF2014} used vehicle location data in time. We experimented with these approaches by adding:
\begin{itemize}
    \item current traffic volume between stops,
    \item weather,
    \item number of buses on a section between stops,
    \item information whether the distance between stops includes a bus lane,
    \item number of intersections with traffic lights,
    \item schedule data.
\end{itemize}


Multiple architectures, accepting bus location and different subsets of available additional data have been tested. For each model, the architecture that gave best results for a given set of experiments, is reported. They are compared to the 
architectures that use only bus location data. 

Travel time distributions differs between days of the week (primarily between working days and weekends). Therefore, following~\cite{ErlendDahl2014}, models were trained on data from the same day, from three consecutive weeks, and tested on data from the last week. 
%
%
Finally, following work reported in Section\ref{SotA}, in the experiments, the accuracy was measured using MAE and STD.

\section{Experimental results and their analysis}\label{experiments}
Let us now summarize experimental results. They are reported representing the ``process of discovery'' that took place in our work. 

{\textbf{Total travel time prediction}} Here, two methods have been tried: ``recursive'' and ``long distance''. In both: number of the next stop, travel times between last three stops, current percentage of the distance traveled between stops, number of the destination stop, current time, route direction, were used. Applied models returned total travel time to the destination.

In the \textit{recursive} method, the model is trained on a data set containing information about travel time to the nearest stop (i.e. focus on prediction of travel time between neighboring stops). Estimating travel time from stop $n$ to stop $n + k$ consists of predicting $k$-steps using the trained model. Specifically, the result from the previous stop is included in the input data for the next prediction. 

The \textit{long distance} method is based on prediction of travel time to a specific stop. Here, the model is trained on data containing information about total pertinent travel time. The assumption was that this approach would be worse for short distances, but may be better at predicting travel time of long(er) trips.

The comparison was made for the bus line 523. The training set consisted of data from Thursdays: March 11, 18, 25. The testing data was from Thursday, April 1. The number of records in the training set was over 1 million, and in the test set over 300,000. Because the goal of this experiment was to compare the methods, not the models, both approaches used the same neural network architecture, i.e. MLP with two hidden layers consisting of 6 and 24 neurons, with ReLU activation function. results are presented in Tables~\ref{tab:CompareRecursiveMethod} and~\ref{tab:CompareLongDistanceMethod}.

\begin{table*}[htbp]
\caption{Prediction of total travel time using the recursive method (bus line 523)}
\label{tab:CompareRecursiveMethod}
\centering
\resizebox{0.75\textwidth}{!}{%
\begin{tabular}{|l|l|l|l|l|l|l|l|l|}
\hline
\multirow{2}{*}{\diagbox{distance}{time}} & \multicolumn{2}{l|}{7:00 - 10:00} & \multicolumn{2}{l|}{10:00 - 14:00} & \multicolumn{2}{l|}{14:00 - 19:00} & \multicolumn{2}{l|}{19:00 - 23:00} \\ \cline{2-9} 
                                &MAE          & STD          & MAE          & STD          &MAE              &STD              &MAE              &STD               \\ \hline
       1 - 3                    &58.24s &62.76s & 60.01s	&79.94s & 59.51s	& 75.74s & 47.02s &	54.01s        \\ \hline
       4 - 8                    & 125.52s	&129.44s & 130.89s	&141.44s & 131.74s	&124.96s & 92.10s	&88.65s      \\ \hline
       9 - 15                    & 211.36s	&165.19s & 214.10s	&179.41s & 240.66s	&170.14s & 156.74s	&121.10s        \\ \hline
       16 - 20                   & 294.83s &	188.44s & 310.42s &	219.57s & 344.49s	&201.74s & 222.66s	&137.57s        \\ \hline
       21 - 27                    & 360.19s &	200.56s & 398.48s	&266.61s & 407.73s &	217.13s & 277.52s &	154.40s      \\ \hline
\end{tabular}%
}
\end{table*}

\begin{table*}[htbp]
\caption{Prediction of total travel time using the long distance method (bus line 523)}
\label{tab:CompareLongDistanceMethod}
\centering
\resizebox{0.75\textwidth}{!}{%
\begin{tabular}{|l|l|l|l|l|l|l|l|l|}
\hline
\multirow{2}{*}{\diagbox{distance}{time}} & \multicolumn{2}{l|}{7:00 - 10:00} & \multicolumn{2}{l|}{10:00 - 14:00} & \multicolumn{2}{l|}{14:00 - 19:00} & \multicolumn{2}{l|}{19:00 - 23:00} \\ \cline{2-9} 
                                &MAE          & STD          & MAE          & STD          &MAE              &STD              &MAE              &STD               \\ \hline
       1 - 3                    &56.72s &	66.06s & 59.97s &	69.67s & 59.07s &	64.44s & 54.82s	&51.11s         \\ \hline
       4 - 8                    & 81.07s&	97.31s& 85.59s&	106.85s & 78.52s &	87.18s & 66.44s	&68.93s      \\ \hline
       9 - 15                   &107.01s &	121.46s & 101.93s&	129.62s & 105.89s &	113.19s & 85.65s&	87.78s        \\ \hline
       16 - 20                   & 130.41s &	144.66s& 140.38s &	170.57s& 130.76s&	138.80s & 113.58s	&106.45s        \\ \hline
       21 - 27                    & 153.70s&	161.04s & 170.72s&	205.03s& 155.09s&	152.63s & 172.30s&	135.29s      \\ \hline
\end{tabular}%
}
\end{table*}

As can be seen, from Tables \ref{tab:CompareRecursiveMethod}  and \ref{tab:CompareLongDistanceMethod}, for all distances longer than 4 stops, the long distance method was more accurate than the recursive method. Moreover, for distances longer than 8 stops, in most time intervals, the results were more than twice as accurate. For 1 to 3 stops, results of both methods were comparable.

\textbf{Architecture comparison} The goal of next experiment was to compare effectiveness of different \textit{RBFN} and \textit{MLP} architects. 
Here, bus lines were considered in eight groups with different characteristics (see, Section~\ref{data}). The goal was to to establish which method is best (MAE), and which is most stable (STD). Overall, five MLP architectures and four RBFN architectures were applied. 

Training days included the same working days of the week from three consecutive weeks (e.g. Wednesdays March, 10, 17 and 24). The test data was from the following week (March 29 to April 2). In this way, five models were created for a single bus line for each architecture. Only data from working days was used, due to the difference in weekend schedules. Moreover, weekends have different traffic patterns, and we wanted to exclude this factor. In total, 29 bus lines and 9 different architectures were tried, which gives a total of 1305 models (29 lines * 9 architectures * 5 days).

The RBFN architectures were implemented in Python using the Keras library, and the RBF~\cite{rbfLayer} code. The hidden layer used Gaussian radial basis function: $exp (-\beta r^2) $. The tested architectures have $M = 10, 15, 25, 35$ neurons in the hidden layer (denoted as RBFN M). The results are presented in Table~\ref{tab:CompareRbfnArchitectures}.
\begin{table*}[htbp]
\caption{RBFN-based prediction results}
\label{tab:CompareRbfnArchitectures}
\centering
\resizebox{0.8\textwidth}{!}{%
\begin{tabular}{|l|l|l|l|l|l|l|l|l|}
\hline
\multirow{2}{*}{\diagbox{group}{model}} & \multicolumn{2}{l|}{RBFN 10} & \multicolumn{2}{l|}{RBFN 15} & \multicolumn{2}{l|}{RBFN 25} & \multicolumn{2}{l|}{RBFN 35} \\ \cline{2-9} 
                                &MAE          & STD          & MAE          & STD          &MAE              &STD              &MAE              &STD               \\ \hline
       Center--Praga                   &\textbf{83.58s} &\textbf{104.18s} &93.28s &114.87s &103.77s &134.16s &121.55s &143.93s          \\ \hline
       Center--periphery                &114.46s &145.65s &115.23s &154.1s &\textbf{103.76s} &139.74s &104.87s &\textbf{135.96s}    \\ \hline
       Long with in periphery North-South     &122.82s &162.4s &108.89s &146.75s &\textbf{102.91s} &\textbf{140.21s} &111.14s &149.3s         \\ \hline
       Long with in periphery East--West   &118.28s &159.77s &123.06s &162.73s &113.49s &152.23s &\textbf{109.46s} &\textbf{149.66s}        \\ \hline
       Long throught the centre               &98.13s &127.49s &100.1s &127.63s &\textbf{95.91s} &122.0s &100.29s &\textbf{117.7s}       \\ \hline
       Express                       &102.01s &133.38s &\textbf{100.02s} &\textbf{125.26s} &100.84s &130.55s &105.04s &134.87s       \\ \hline
       Short with in center               &94.72s &118.68s &\textbf{86.0s} &\textbf{109.44s} &87.33s &111.41s &103.81s &132.3s       \\ \hline
       Short with in periphery               &\textbf{89.07s} &116.0s &89.17s &\textbf{115.54s} &89.32s &115.87s &102.78s &132.06s       \\ \hline
       All groups                   & 102.88s &133.44s & 101.96s	&132.04s & \textbf{99.66s}  &\textbf{130.77s} & 107.36s &136.97s     \\ \hline
\end{tabular}%
}
\end{table*}

It can be seen that for short routes (\textit{Center--Praga, short within Center, short within periphery}) and \textit{Express} routes, the most accurate results were obtained by the ``smaller'' RBFN architectures (RBFN 10 and RBFN 15). For the remaining routes, RBFN 25 and RBFN 35 performed better. This is likely because, for long paths, more diverse data ``needs more neurons'' in the hidden layer.

Overall, the RBFN 25 seems to be the ``optimal architecture''. Only, for \textit{Express}, \textit{Short within center}, and \textit{Short within periphery}, its slightly worse than RBFN 10 and RBFN 15. The difference in MAE was between 0.25s and 1.33s, and in STD between 0.33s and 5.29s, which is not significant.

For MLP (implemented in Python with Keras) five architectures were tried: networks with 2 or 3 hidden layers and: [6, 12], [12, 32], [6, 8, 12], [8, 8] neurons. Here, ReLU and tanh were compared as activation functions. The results are presented in Table~\ref{tab:CompareMlpArchitectures}. 
\begin{table*}[htbp]
\caption{MLP-based prediction results}
\label{tab:CompareMlpArchitectures}
\centering
\resizebox{\textwidth}{!}{%
\begin{tabular}{|l|l|l|l|l|l|l|l|l|l|l|}
\hline
\multirow{2}{*}{\diagbox{grupa}{model}} & \multicolumn{2}{l|}{MLP 6, 12 ReLU} & \multicolumn{2}{l|}{MLP 12, 32 ReLU} & \multicolumn{2}{l|}{MLP 12, 32 ReLU tanh} & \multicolumn{2}{l|}{MLP 6, 8, 12 ReLU} & \multicolumn{2}{l|}{MLP 8, 8 tanh} \\ \cline{2-11} 
                                &MAE          & STD          & MAE          & STD          &MAE              &STD              &MAE              &STD    &MAE              &STD             \\ \hline
       Center--Praga                   &112.33s &143.9s &96.43s &119.94s &\textbf{94.08s} &\textbf{118.30s} &114.5s &147.78s &115.93s &146.75s     \\ \hline
       Center--periphery                 &\textbf{101.54s} &\textbf{133.53s} &106.9s &135.26s &110.86s &146.74s &110.69s &145.8s &143.31s &180.73s     \\ \hline
       Long within periphery North-South      &139.06s &178.95s &119.79s &\textbf{159.51s} &\textbf{114.39s} &164.76s &120.05s &164.01s &136.25s &173.61s       \\ \hline
       Long within periphery East--West   &122.47s &167.79s &\textbf{113.67s} &\textbf{151.78s} &133.6s &181.72s &124.23s &162.63s &154.59s &196.67s     \\ \hline
       Long through the centre               &109.35s &146.14s &\textbf{97.71s} &\textbf{128.49s} &118.17s &153.16s &105.89s &139.59s &109.16s &143.21s     \\ \hline
       Express                        &127.52s &168.45s &\textbf{92.46s} &\textbf{123.74s} &118.71s &157.53s &116.61s &154.02s &132.97s &174.49s    \\ \hline
       Short within center                 &97.02s &130.73s &\textbf{95.42s} &\textbf{121.13s} &108.07s &139.06s &106.31s &132.19s &101.55s &135.38s    \\ \hline
       Short within periphery               &107.13s &133.08s &\textbf{90.24s} &\textbf{122.65s} &96.44s &124.07s &99.89s &129.8s &104.84s &139.98s     \\ \hline
       All groups                   & 114.55s &150.32s & \textbf{101.57s}	&\textbf{132.81s} & 111.78s  &148.16s & 112.27s &146.97s &124.82s &161.35s      \\ \hline
\end{tabular}%
}
\end{table*}

Overall, the most effective and stable architecture for all groups was two hidden layers consisting of 12 and 32 neurons, with ReLU as the activation function. Note, however, that the results are somewhat more complex. The network with the tanh activation function in the last layer, provided slightly better results for the \textit{Center--Praga} lines. Moreover, for the \textit{Long on the north-south periphery} lines it had value of MAE lower by 5.40s, but higher STD value by 5.25s. On the other hand, the [6, 12] ReLU network, for the \textit{Center--periphery} lines, had MAE lower by 5.36s, and STD lower by 1.73s. Overall, the [12, 32] ReLU network can be seen as the ``winner''. Therefore it was used in subsequent experiments. Finally, the MLP using only tanh achieved worst results. 

The next step was to compare RBFN 25 and MLP [12,32] ReLU. Based on the travel time ratio (\textit{average travel time in a given hour}/\textit{shortest travel time during the day}), two rush hour periods have been established: (1) the morning peak -- 7:00 am to 10:00 am, (2) the afternoon peak -- 3:00 pm to 7:00 pm. During peak hours, the predictive models encounter problems to correctly determine travel time. Therefore, a detailed comparison of the results of RBFN 25 and MLP [12, 32] ReLU was performed for the ``worst-case conditions''.

Based on the analysis of obtained results it has been concluded that: (a)~on short distances in the morning peak, both models performed quite similarly. Only in the case (b)~\textit{Long at the the north-south periphery} lines, the effectiveness of the RBFN was better than the MLP. On the other hand, for (c)~\textit{Express} routes, a significant advantage of the MLP was observed. For (d)~distances ranging from 6 to 15 stops, models are comparable, with minor differences for \textit{Center -- Praga} and \textit{Short in the periphery}, lines, where the MLP has advantage. In the case of (e)~long distances, from 16 to 25 stops, for the \textit{Express} lines, the RBFN model significantly exceeded the MLP. Interestingly, (f)~for the same group, the MLP performed significantly better over short distances. For the (g)~remaining long-distance groups, models achieved similar performance, with a slight advantage of the RBFN for the four groups and a slight advantage of the MLP for the other three groups. Similar tendencies have been observed (h)~for predictions during the afternoon peak. Interestingly, (i)~performance of models in the afternoon peak was better than in the morning peak. This is probably due to the fact that there are more buses on the route in the afternoon than in the morning hours, so the training set contains more data. Hence, models could better capture specificity of travel time during afternoon hours.

Overall, forecasting travel time, based only on the location data, the RBFN and MLP provided similarly accurate forecasts, with a slight advantage of RBFN. Note, however, that for long distances from 16 to 25 stops, in the morning and afternoon rush hours, RBFN achieved significantly better prediction accuracy for 3 groups, slightly better accuracy for 8 groups, and for two groups, the accuracy of both models was almost equal. The MLP, for this range, was much better only for the \textit{Center--Praga} lines in the afternoon, and slightly better for the \textit{Center--periphery} and the  \textit{Center--Praga} lines in the morning peak.

Separately, consider accuracy of the predictions of the best models, for each bus group, in the most difficult conditions, i.e. over long distances during rush hours. For 6 out of 8 groups, the accuracy of approximately 75\% of predictions had an error of less than 200s. Only for the \textit{Express} and \textit{Long within periphery West-East} lines, the accuracy of about 75\% of the predictions had an error of less than 260s. Material presented in what follows aims at improving the results obtained using only location information.

Next experiment used Historical Averages method (HA,~\cite{WeiFan,Zychowski2017}), which applies average travel times from previous days to estimate the current travel time. Here, prediction depends on days of the week and rush hours. Hence, averages are calculated separately for working days and weekends. Based on the GPS data, the departure time from the last visited stop can be determined. Then, based on historical data, the estimated time of arrival at the next stops is determined. For each bus line, the data from the training sets (i.e. all working days from March,8 to 26) was divided into 20-minute groups, i.e. records sent every day between 10:00-10:20 belonged to one group. The average travel times between current locations, and all stops to the end of the route were calculated. For example, if for time period 10:00-10:20 data included 10 buses located on the 6-th stop of the line, the average times of travel this stop to all stops, till the end of their lines, were calculated for these 10 buses. In this way, the HA algorithm approximated the estimated travel time between any stops on the route for any given hour. The average travel times, determined using this algorithm, have been stored as the training sets. Then, for each bus line, for all test sets (i.e. data for working days from March,29 to April,2), travel time predictions were calculated using the HA algorithm. 
In addition, analogous calculations were carried out while the average travel times were calculated for the same day of the week only. For example, travel time predictions for March 31 (Wednesday) were based on data data from March 10, 17, 24. Table~\ref{tab:time_table_vs_ha_means} presents the MAE values of travel time predictions using HA method. 
\begin{table}[htbp]
\caption{Prediction MAE for timetable, HA for same days, HA for working days.}
\label{tab:time_table_vs_ha_means}
\centering
\resizebox{0.75\textwidth}{!}{
\begin{tabular}{|c|c|c|c|}
\hline
  group & time table  & HA(the same day ) & HA(all days ) \\ \hline
Long within periphery North-South & 79.22s          & 81.53s   & 71.94s  \\                              \hline
Long within periphery West-East  & 80.85s          & 75.02s                                       & 62.38s                                      \\ \hline
Center--periphery                    & 64.76s          & 62.02s                                       & 61.62s                                      \\ \hline
Long through the centre                 & 71.58s          & 68.75s                                       & 67.64s                                      \\ \hline
Express                           & 60.38s          & 64.65s                                       & 52.70s                                      \\ \hline
Center--Praga                      & 64.56s          & 52.89s                                       & 50.22s                                      \\ \hline
Short within periphery                  & 55.87s          & 50.92s                                       & 35.65s                                      \\ \hline
Short within centre                   & 60.41s          & 53.02s                                       & 49.02s                                      \\ \hline
\end{tabular}%
}
\end{table}

\begin{figure*}[ht]
  \centering
   \includegraphics[width=1.0\textwidth]{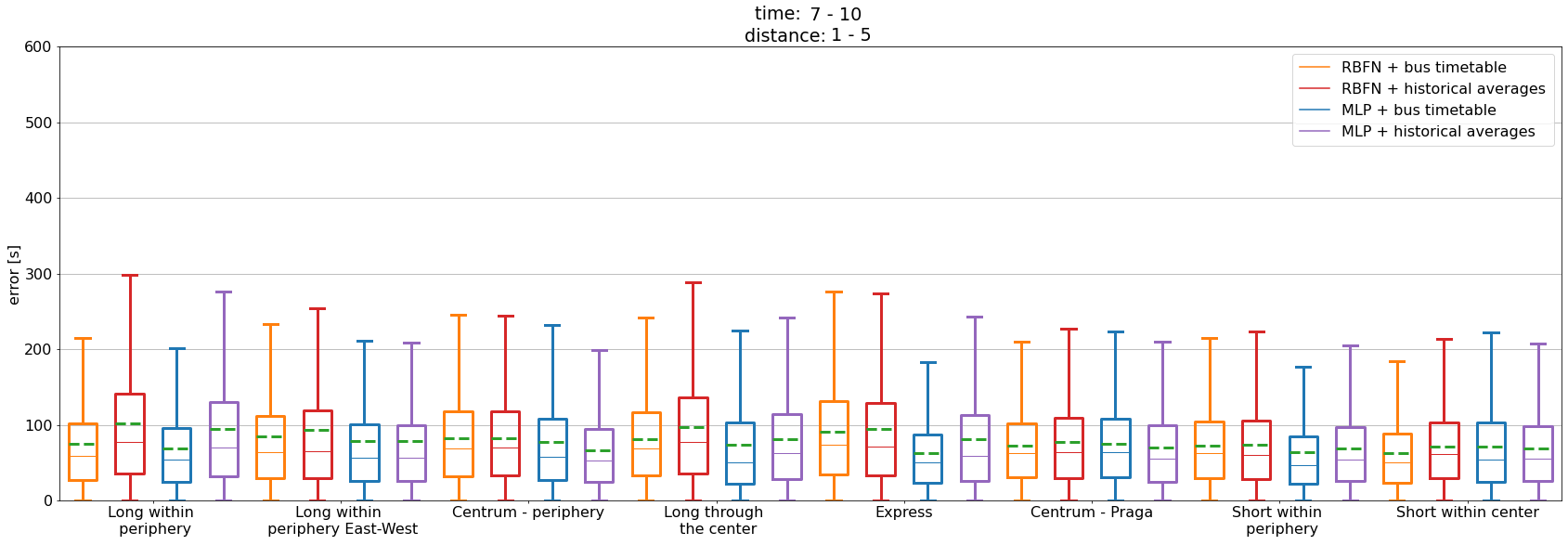}\hfill
    \\[\smallskipamount]
    \includegraphics[width=1.0\textwidth]{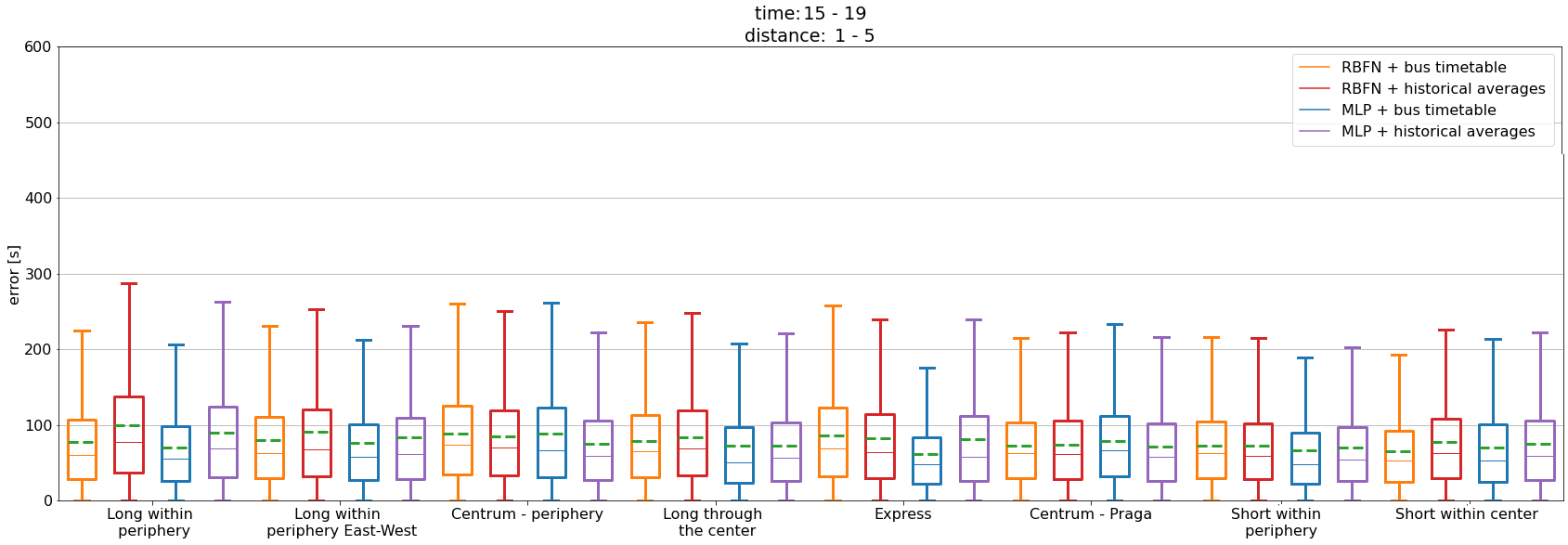}\hfill
    \caption{Comparison of the prediction accuracy of the RBFN and MLP models with the additional features of the predicted scheduled travel time and delay, and of the models with the additional feature of the predicted travel time determined by the HA algorithm. Results include predictions for distances from 1 to 5 stops during rush hour.}
  \label{fig:table_time_vs_hist_aver_1_5}
\end{figure*}

\begin{figure*}[ht]
    \centering
    \includegraphics[width=1.0\textwidth]{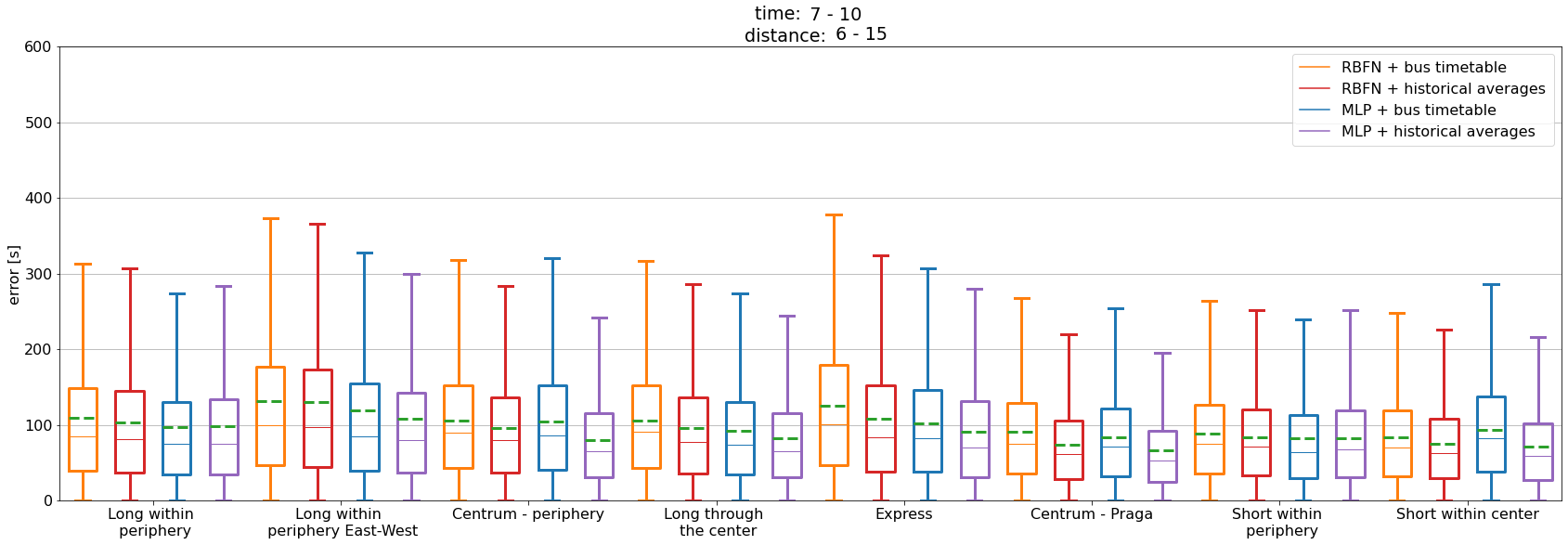}\hfill
   \\[\smallskipamount]
   \includegraphics[width=1.0\textwidth]{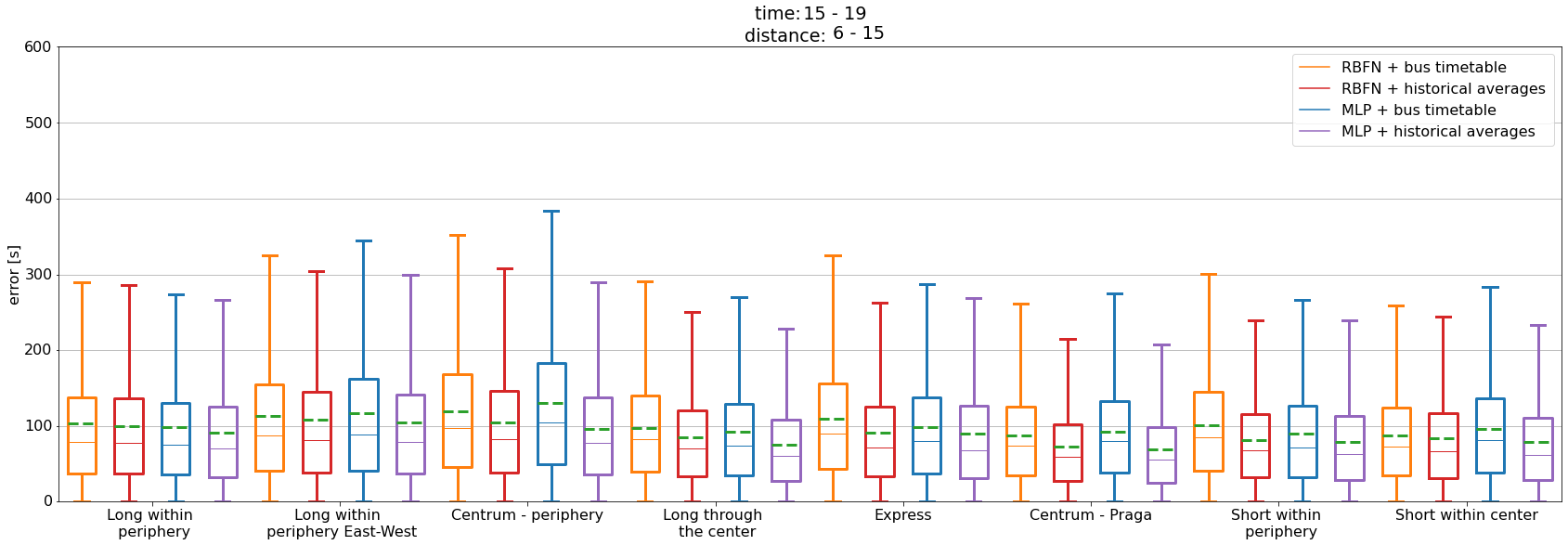}\hfill
   \caption{Comparison of the prediction accuracy of the RBFN and MLP models with the additional features of the predicted scheduled travel time and delay, and of the models with the additional feature of the predicted travel time determined by the HA algorithm. Results include predictions for distances from 6 to 15 stops during rush hour.}
   \label{fig:table_time_vs_hist_aver_6_15}
\end{figure*}

\begin{figure*}[ht]
    \centering
    \includegraphics[width=1.0\textwidth]{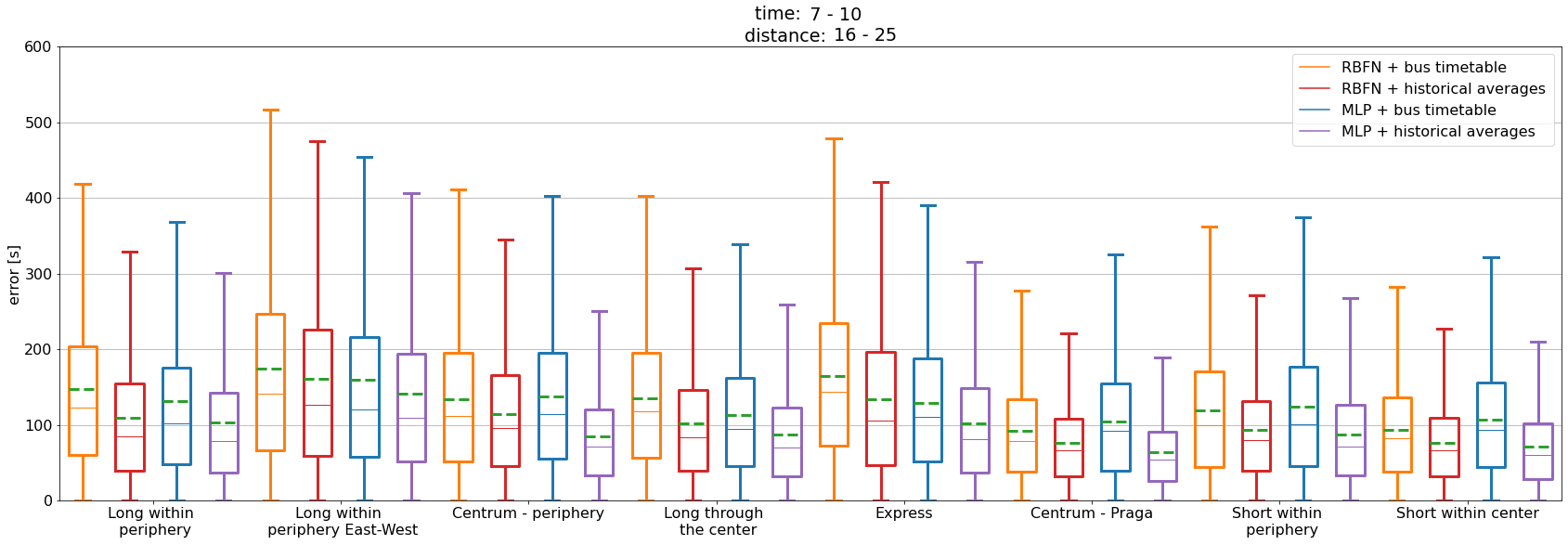}\hfill
    \\[\smallskipamount]
    \includegraphics[width=1.0\textwidth]{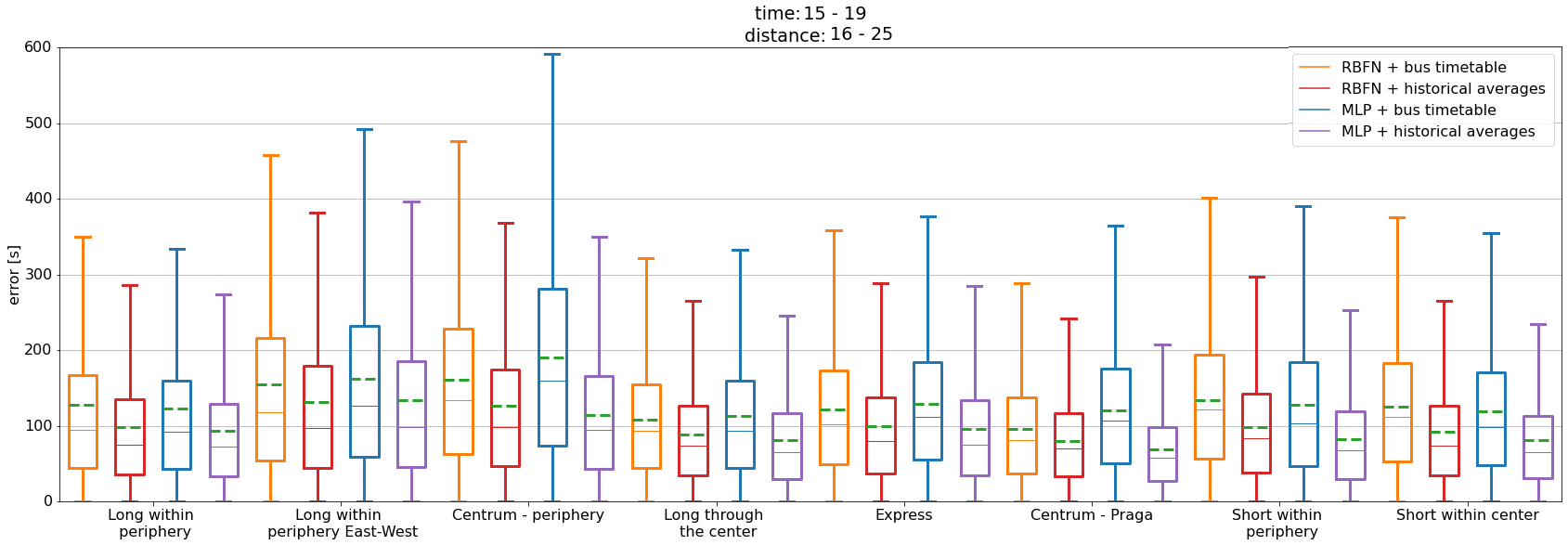}\hfill
    \caption{Comparison of the prediction accuracy of the RBFN and MLP models with the additional features of the predicted scheduled travel time and delay, and of the models with the additional feature of the predicted travel time determined by the HA algorithm. Results include predictions for distances from 16 to 25 stops during rush hour.}
   \label{fig:table_time_vs_hist_aver_16_25}
\end{figure*}

As can be seen, HA based on data from all working days provides better accuracy then when using data from the same day of the week. Moreover, observe significant improvement in the accuracy of prediction using the HA algorithm, compared use of travel time found in the timetables. Based on these observations, the idea was born to replace the \textit{travel time predicted by distribution data} with \textit{travel time predicted by the HA algorithm} in the NN-based travel time prediction. This led to creation of a hybrid model. Here, averages for all working days were used. The determined estimated travel times replaced the \textit{estimated travel time from the timetable data} feature, and the \textit{last stop delay} feature in the predictions with MLP and RBFN. 
The graphs~\ref{fig:table_time_vs_hist_aver_1_5}, \ref{fig:table_time_vs_hist_aver_6_15}, \ref{fig:table_time_vs_hist_aver_16_25} 

Comparison of experimental prediction accuracy of RBFN and MLP models, using HA and not, brought the following conclusions. At distances from 1 to 5 stops, models using distribution data were more accurate than predictions using HA, for 6 of 8 groups: \textit{Long within periphery North--South} (RBFN and MLP), \textit{Long within periphery East--West} (RBFN and MLP), \textit{Long within the center} (RBFN and MLP), \textit{Express} (MLP), \textit{Short within periphery} (MLP), \textit{Short within the center} (RBFN). For the remaining groups, the prediction errors were comparable. The only exception was the MLP in the \textit{Center-periphery} group, where the results were more accurate when using HA data. On the other hand, for distances from 6 to 15 stops, in the chart~\ref{fig:table_time_vs_hist_aver_6_15} 
better results when using HA, were established for \textit{all} groups. The greatest improvement, when using HA, was noted for predictions for long distances from 15 to 26 stops. 
In the ~\ref{fig:table_time_vs_hist_aver_16_25} chart, you can see a significant reduction in the prediction error of models using HA compared to models using distribution data. 
Here, for \textit{Center--periphery}, \textit{Center--Praga}, \textit{Short within periphery}, \textit{Short within the center} lines, for MLP, MAE decreased almost twice. For the RBFN, drops of MAE were also significant, reaching up to 30s, for \textit{Long within periphery North-South} lines.

Based on these results it has been concluded that for 1 to 5 stops, RBFN or MLP models, using the \textit{estimated travel time} feature, calculated on the basis of data from timetables, work better. However, models using HA-based data, deliver slightly better performance for 6 to 15 stops. In the case of predictions for 16 to 25 stops, the use of HA significantly reduces the prediction error.

This suggests a ``hybrid model'' for prediction of bus travel time. Hence, in Table~\ref{tab:summapry_best_models}, information, which models provide best prediction accuracy, depending on the group of bus lines, and the travel distances is summarized.
\begin{table*}[htbp]
\caption{Models with best travel time prediction by distance and group of bus lines.}
\label{tab:summapry_best_models}
\centering
\resizebox{0.75\textwidth}{!}{%
\begin{tabular}{|l|l|l|l|}
\hline
\diagbox{group}{distance}  & up to 5 stops  & 6 to 15 stops     & more than 16 stops   \\ \hline
Long within periphery & MLP + table time  & MLP + HA  & MLP + HA \\ \hline
Long within periphery East-West   & MLP + time table  & MLP + HA  & MLP + HA \\ \hline
Center--periphery & MLP + table time  & MLP + HA  & MLP + HA \\ \hline
Long through the center  & MLP + time table  & MLP + HA  & MLP + HA \\ \hline
Express  & MLP + time table  & MLP + HA  & MLP + HA \\ \hline
Center--Praga   & RBFN + time table & MLP + HA  & MLP + HA \\ \hline
Short within periphery   & MLP + time table  & RBFN + HA & MLP + HA \\ \hline
Short within center    & RBFN + time table & MLP + HA  & MLP + HA \\ \hline
\end{tabular}
}
\end{table*}

Based on results from Table~\ref{tab:summapry_best_models}, a hybrid model was developed, consisting of: RBFN 25 or MLP [12, 32] with the features: predicted travel time based on schedule data and delay at the last stop, and RBFN 25 or MLP [12, 32] with estimated travel time using HA. When making predictions, depending to which group given bus line belongs, and the distance for which the prediction is performed, the hybrid approach uses models specified in Table~\ref{tab:summapry_best_models}.

\begin{figure*}[htbp]
    \centering
    \includegraphics[width=.4\textwidth]{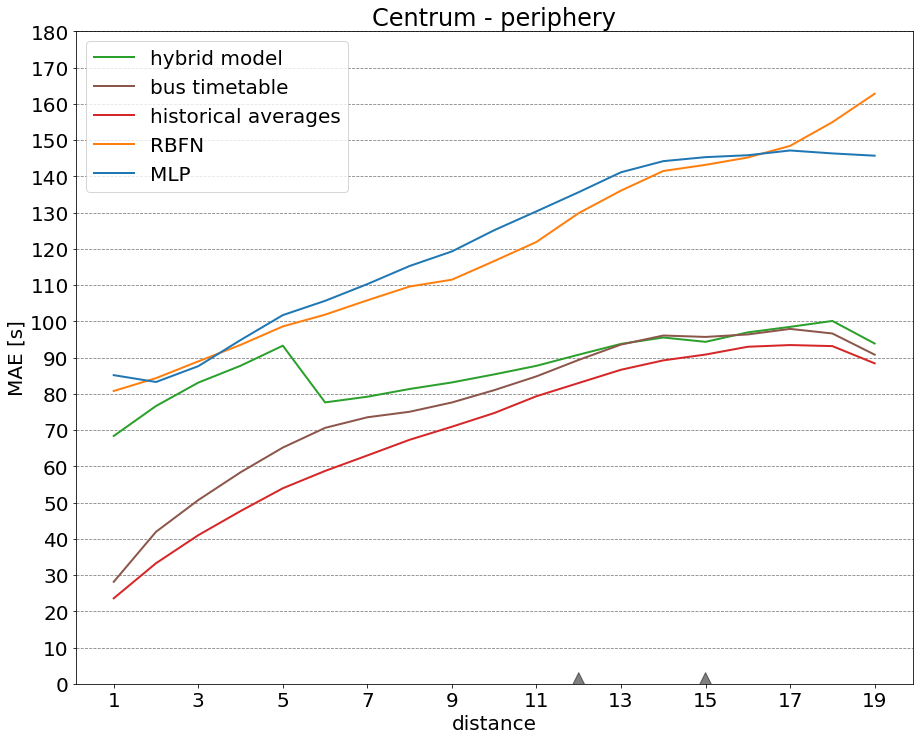}\hfill
    \includegraphics[width=.45\textwidth]{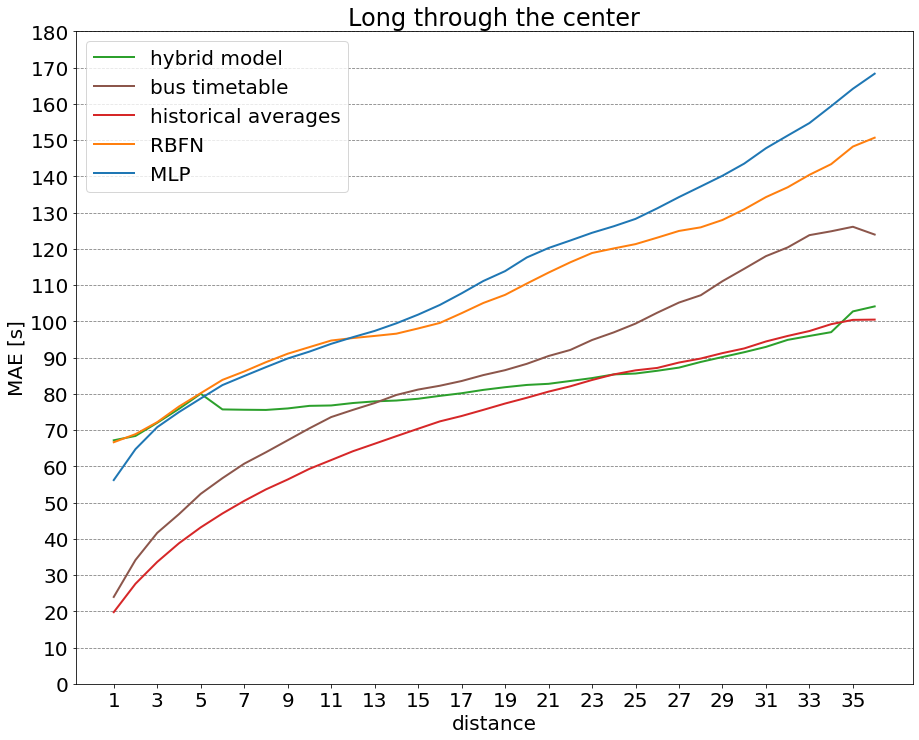}\hfill
    \\[\smallskipamount]
   \includegraphics[width=.45\textwidth]{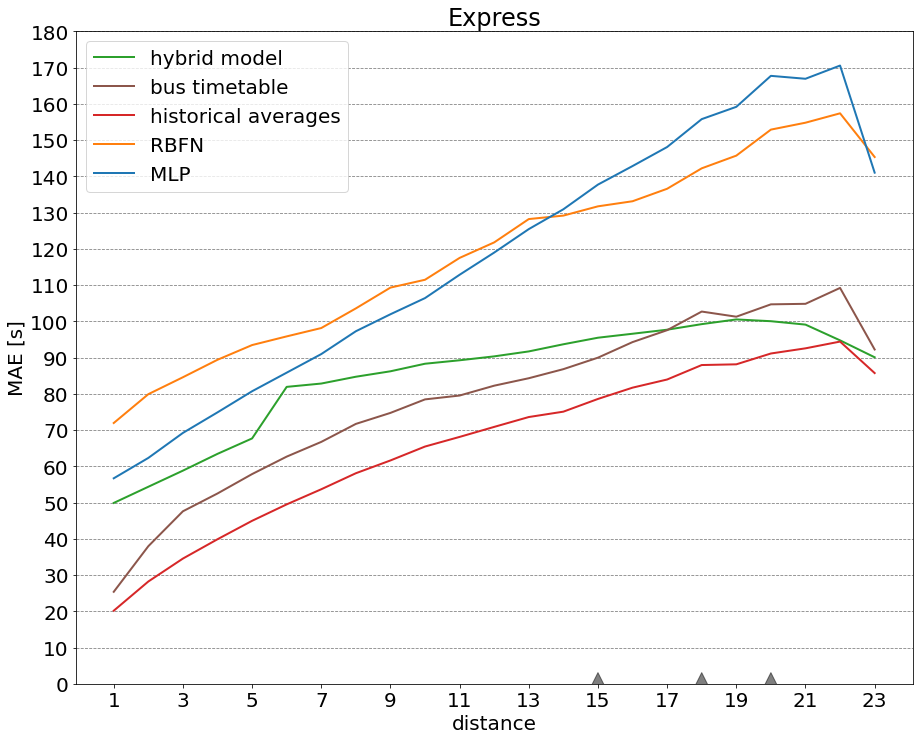}\hfill
    \includegraphics[width=.45\textwidth]{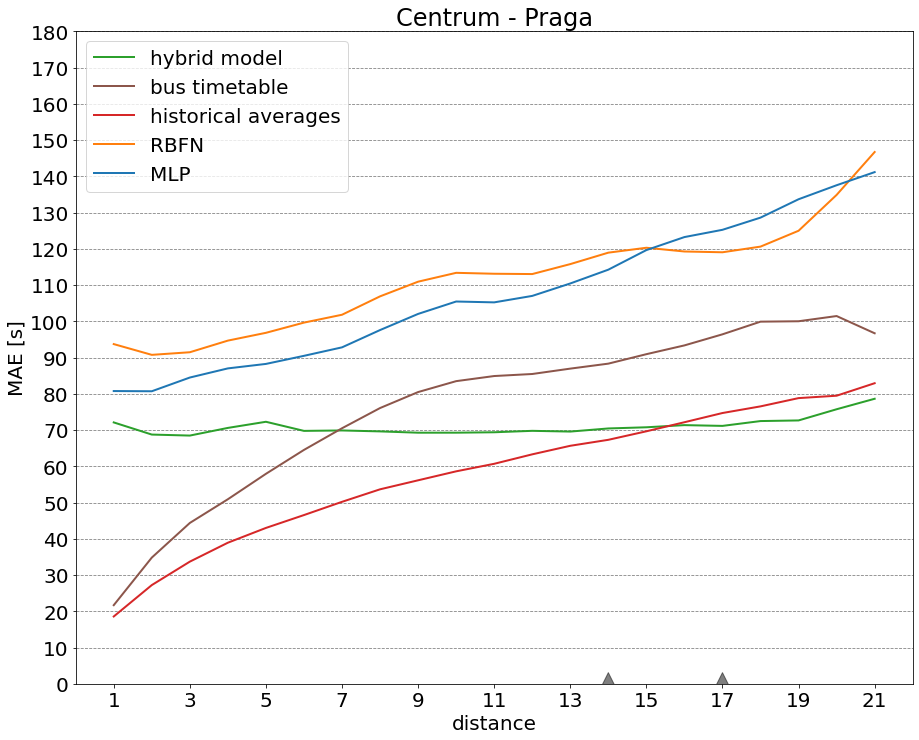}\hfill
   \\[\smallskipamount]
    \includegraphics[width=.4\textwidth]{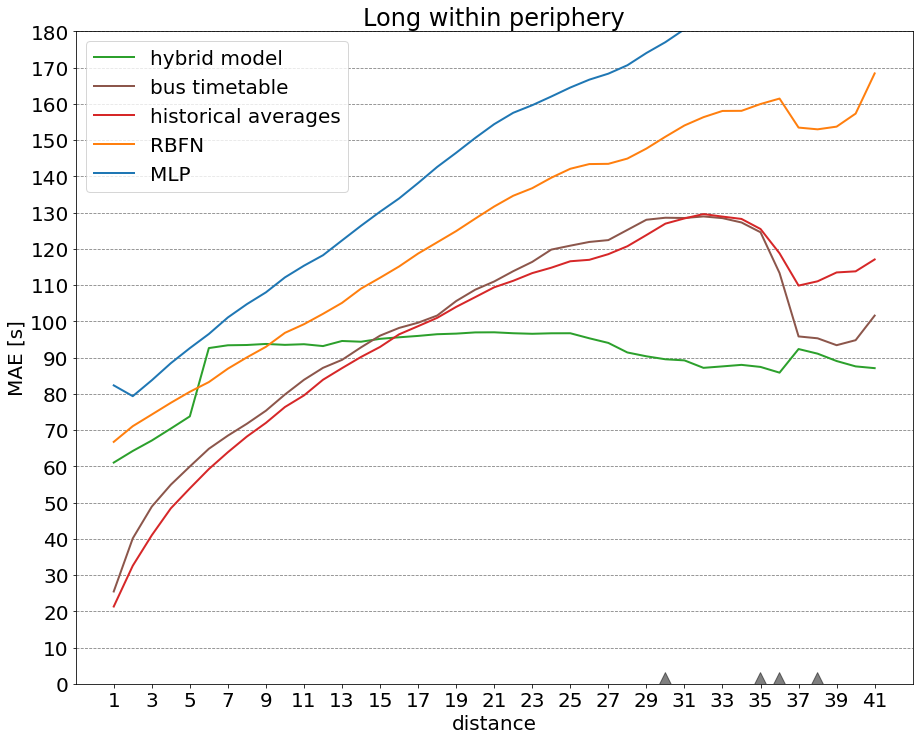}\hfill
    \includegraphics[width=.45\textwidth]{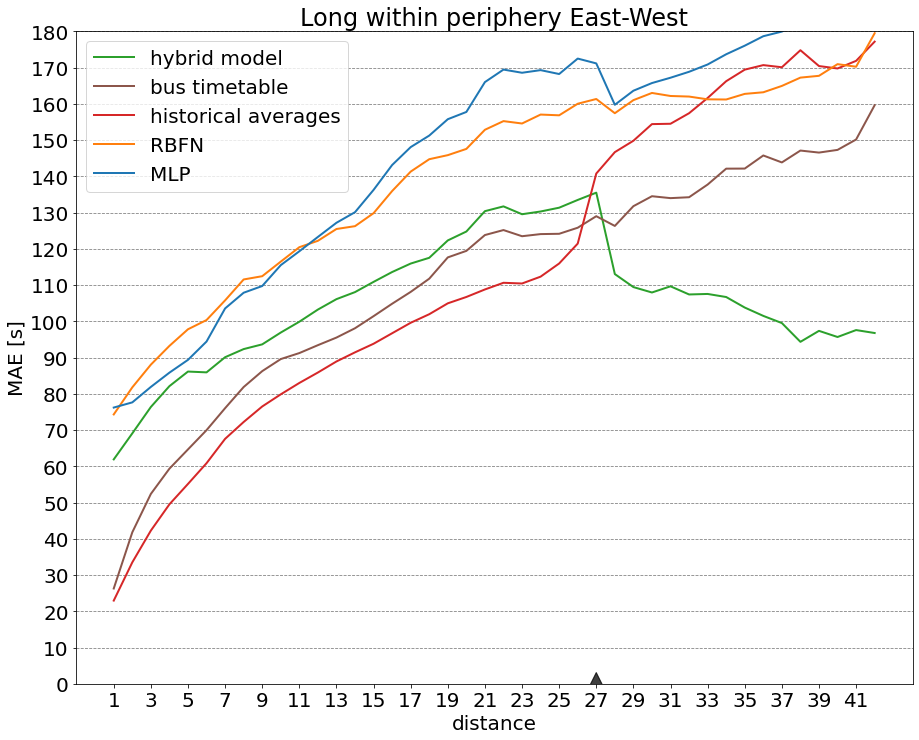}\hfill
    \\[\smallskipamount]
    \includegraphics[width=.45\textwidth]{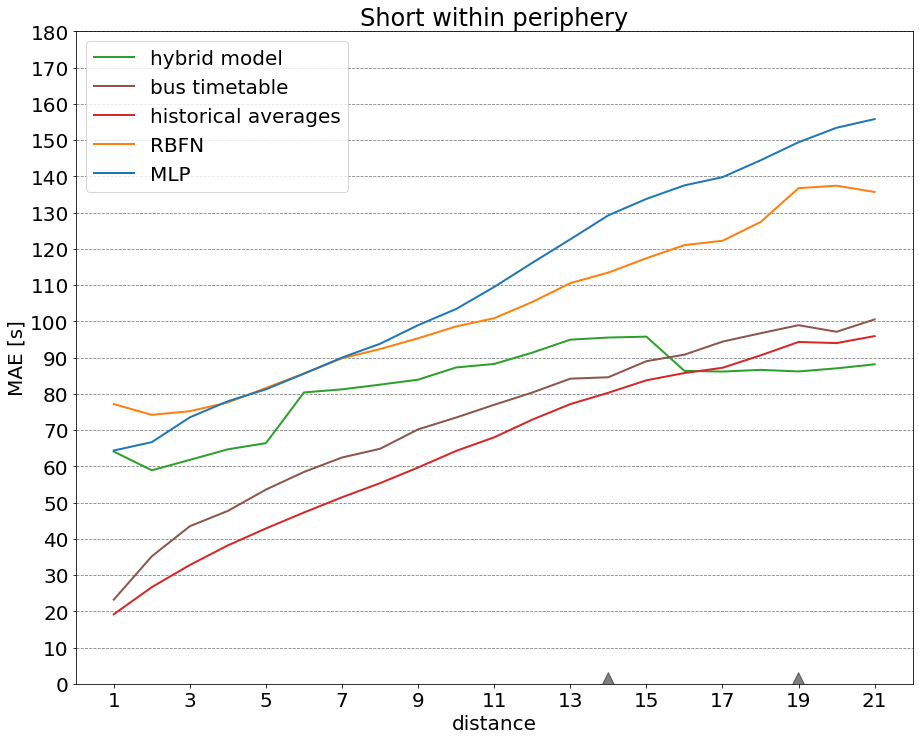}\hfill
    \includegraphics[width=.45\textwidth]{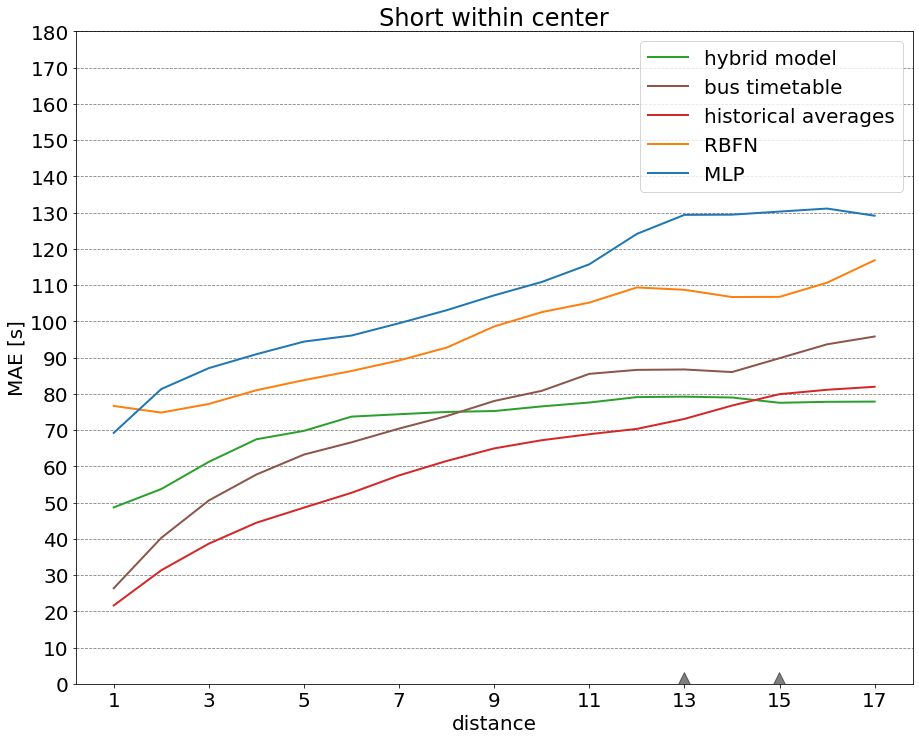}\hfill
    \caption{Comparison of the MAE value of prediction methods at different distances}
    \label{fig:summary_mae_groups}
\end{figure*}

The created hybrid model was compared with: MLP and RBFN models using a basic set of features, the HA algorithm, and predictions based on data from timetables. Figure~\ref{fig:summary_mae_groups} represents comparison of MAE values (in seconds) of the predictions made by the created hybrid model with the previously mentioned methods, depending on the distance (number of stops) for which the prediction was made. Due to space limitation, only two, distinct, cases are reported. 
Bus lines belonging to the same group had different lengths of routes. Therefore, black triangles mark distances for which the prediction of travel time ends for one of bus lines belonging to the group described by the graph. For this reason, the graphs report rapid changes in MAE for adjacent distances, as in the case of \textit{Long within periphery} group, where lines have total of 30, 35, 36 and 38 stops. 

Results represented in figures, and the remaining experimental results, show that for short-distance predictions, for all groups, the HA algorithm and distribution times delivered much better accuracy than the hybrid model. However, for the groups \textit{Short within center}, \textit{Short within periphery}, \textit{Long within periphery West East}, \textit{Long within periphery North-South}, \textit{Center-Prague}, for long distances (from 15 stops for \textit{Short within center}, from 17 stops for \textit{Short within periphery}, from 27 stops for \textit{Long within periphery West East}, from 16 stops for \textit{Long within periphery North-South} group, and from 16 stops for \textit{Center-Prague} group), the hybrid model was most accurate. 

More specifically, for the \textit{Long within periphery North-South} group, the hybrid model provides the best accuracy for as much as 60\% of distances. In the group \textit{Long within periphery West-East}, for travel time predictions from 27 stops, where the predictions were made only for line 102 (only line this long), the MAE of the hybrid model was 95-110s, while the MAE of the second best method (HA), was 130-150s. In the case of the \textit{Center-periphery} group, for distances from 13 stops, the accuracy of the hybrid model was at the level of the time table method. Here, the HA algorithm had the lowest MAE for all distances. In the \textit{Long through center} group, for prediction from 23 stops, the MAE of the hybrid model was similar to the HA algorithm. For the \textit{Express} route group, apart from the travel time predictions for distances greater than 21 stops, the MAE of the hybrid model was much higher than the MAE of the HA algorithm. It is also worth noting that the accuracy of the basic RBFN and MLP was significantly inferior to all other methods that have been experimented with.

\section{Comparison with the models from the SotA}

Let us now report the best MAE results found in the literature, and ``compare'' it with the results reported here. The top 5 lowest MAE values are as follows. (1) ANN from~\cite{ErlendDahl2014}; MAE at 40s; for predictions on distances of up to 16 stops. (2) Hybrid model reported here; MAE at 67.17s; for predictions for the \textit{Short through the center} group, for time interval 19:00 to 23:00. (3) ANN from~\cite{WeiFan}; MAE at 70s. (4) BP from~\cite{BP2019}; MAE at 125s; for prediction for morning rush hours. (5) MLP from~\cite{Zychowski2017}; MAE at 138.40s; for prediction of travel time of \textbf{trams} for noon hours.

However, it has to be stressed that these comparisons represent only a meta level view of possible accuracy of travel time prediction. In each article, results were obtained for bus (or tram) lines from different cities, where each city has very different road and traffic structure and public transport characteristics. For example, in more populated cities, or those with less developed road infrastructure, there may be more delays due to heavy traffic, which may influence the accuracy of model prediction. Moreover, Trondheim (city population is around 200 thousands, while metropolitan population is around 280000) is a much smaller city than others, which may influence prediction accuracy. Further, Warsaw is split by a river, with limited number of bridges, through which all bus lines have to pass to get ``to the other side''. Besides, each of analyzed bus/tram lines had different lengths. Finally, no other work used data for multiple bus lines individually or in groups. Regardless, it can be argued that the results reported in this contribution are very competitive and worthy further explorations.

\section{Concluding remarks}\label{conclusions}

The aim of this work was to extend knowledge concerning applicability of machine learning to prediction of travel times of public transport. The main contributions of this work are related to the systematic study of interrelations between characteristics of groups of bus lines and applied ML algorithms. While it is extremely difficult to fairly compare results obtained for cities located in India, Northern Europe and China, it is clear that the obtained results are very competitive. Moreover, even though there was no space to provide detailed results, performed experiments showed that use of auxiliary features, e.g. number of busses moving between stops, or number of crossings with lights does not lead to important improvement of accuracy of prediction.

Pre-processed data is available at \textit{\url{https://tinyurl.com/mvt4hu3f}}.

\bibliographystyle{splncs04}

\end{document}